\documentclass[10pt,twocolumn,letterpaper]{article}

 \usepackage{cvpr}              
\definecolor{cvprblue}{rgb}{0.21,0.49,0.74}
\usepackage[pagebackref,breaklinks,colorlinks,allcolors=cvprblue]{hyperref}
\usepackage{multirow}
\usepackage{array}
\newcolumntype{M}[1]{>{\centering\arraybackslash}m{#1}}

\def\paperID{} 
\def\confName{CVPR}
\def\confYear{2026}

\title{GS\textasciicircum2: Graph-based Spatial Distribution Optimization for Compact 3D Gaussian Splatting}

\author{
Xianben Yang$^{1}$ \quad  Tao Wang$^{1}$\thanks{Denotes co-corresponding authors.}\quad  Yuxuan Li$^{1}$\quad  Yi Jin$^{1}$\footnotemark[1] \quad  Haibin Ling$^{2}$\\
$^{1}$	Key Laboratory of Big Data and Artificial Intelligence in Transportation, Ministry of Education;\\
	School of Computer and Information Technology, Beijing Jiaotong University, Beijing 100044, China\\
$^{2}$ Department of Artificial Intelligence, Westlake University, Hangzhou 310030, China\\
{\small\ttfamily \{yxb\_2023, twang, 23120376, yjin\}@bjtu.edu.cn\quad  linghaibin@westlake.edu.cn}
}

\begin{document}
\maketitle
\begin{abstract}
3D Gaussian Splatting (3DGS) has demonstrated breakthrough performance in novel view synthesis and real-time rendering.   Nevertheless, its practicality is constrained by the high memory cost due to a huge number of Gaussian points.   Many pruning-based 3DGS variants have been proposed for memory saving, but often compromise spatial consistency and may lead to rendering artifacts. To address this issue, we propose graph-based spatial distribution optimization for compact 3D Gaussian Splatting (GS\textasciicircum2), which enhances reconstruction quality by optimizing the spatial distribution of Gaussian points. Specifically, we introduce an evidence lower bound (ELBO)-based adaptive densification strategy that automatically controls the densification process. In addition, an opacity-aware progressive pruning strategy is proposed to further reduce memory consumption by dynamically removing low-opacity Gaussian points. Furthermore, we propose a graph-based feature encoding module to adjust the spatial distribution via feature-guided point shifting.  Extensive experiments validate that GS\textasciicircum2 achieves a compact Gaussian representation while delivering superior rendering quality. Compared with 3DGS, it achieves higher PSNR with only about 12.5\% Gaussian points. Furthermore, it outperforms all compared baselines in both rendering quality and memory efficiency.  The source code is publicly available at 
\href{https://github.com/BJTU-KD3D/GS-2}{https://github.com/BJTU-KD3D/GS-2}.
\end{abstract}    
\section{Introduction}
\label{sec:intro}

Novel view synthesis is critical in computer vision and computer graphics, and has attracted great attention in many applications, such as augmented reality~\cite{Chen_2025_CVPR} and autonomous driving~\cite{ICLR2025_93b4d708}. \textit{Neural Radiance Fields} (NeRF)~\cite{mildenhall2021NeRF} generates high-fidelity images through volume rendering, demonstrating excellent detail preservation. However, its time-consuming network inference limits its application in practical scenarios. Recently, \textit{3D Gaussian Splatting} (3DGS)~\cite{kerbl20233d} has attracted intensive attention due to its promising rendering quality and computational efficiency. 3DGS enhances detail representation through densification, enabling more efficient and higher-quality 3D scene reconstruction than NeRF.

\begin{figure}[t]
    \centering
    \begin{tabular}{@{}c@{\hspace{0.5mm}}c@{\hspace{1mm}}c@{}}
        \raisebox{0.1\height}{\rotatebox{90}{\textbf{\small Render View}}} &
        \includegraphics[width=0.475\linewidth]{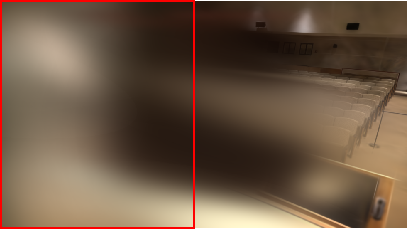} &
        \includegraphics[width=0.475\linewidth]{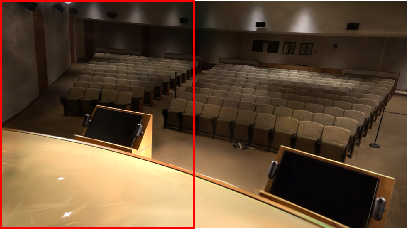} \\[1pt]

        \raisebox{0.45\height}{\rotatebox{90}{\textbf{\small GS View}}} &
        \includegraphics[width=0.475\linewidth]{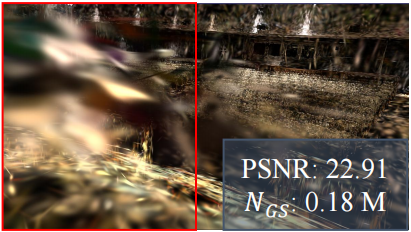} &
        \includegraphics[width=0.475\linewidth]{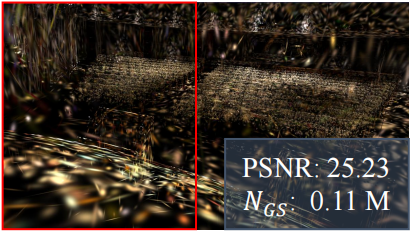} \\[-1pt]

         & \small (a) LightGaussian & \small (b) Ours
    \end{tabular}
    
    \caption{Comparison of rendering quality and spatial distribution. We report the PSNR and \textit{number of Gaussian points} ($N_{\textrm{GS}}$, in millions) for each method. While a SOTA solution (LightGaussian~\cite{fan2024lightgaussian}) compromises spatial consistency with visual artifacts, our method produces a more uniform and coherent distribution.
} \vspace{-4mm}
    \label{fig:teaser}
\end{figure}

While 3DGS effectively densifies sparse \textit{Structure from Motion} (SfM) points into a dense set of Gaussian points, the process often leads to over-parameterization, resulting in excessive memory overhead and reduced rendering speed. For example, unbounded 360-degree scenes from the Mip-NeRF 360 dataset~\cite{barron2022mip} typically require over 3 million Gaussian points when reconstructed using 3DGS~\cite{kerbl20233d}. Many approaches attempt to alleviate this burden by substantially reducing the number of Gaussian points through pruning~\cite{lee2024c3dgs, liu2025maskgaussian, girish2024eagles, fan2024lightgaussian}. However, pruning Gaussian points often disrupts spatial consistency and continuity, resulting in  noticeable rendering artifacts, as illustrated in~\cref {fig:teaser}. These methods ignore the global and local continuity constraints, leading to noticeable degradation in rendering quality, as shown in~\cref{table:baseline1}.

To address these challenges, we introduce \textbf{G}raph-based \textbf{S}patial distribution optimization for compact 3D \textbf{G}aussian \textbf{S}platting (GS\textasciicircum2).  Our method utilizes a graph-based feature encoding network along with global alignment loss and local smoothness loss to optimize the spatial distribution of Gaussian points after pruning. Specifically, GS\textasciicircum2 first optimizes densification to avoid redundant Gaussian points overgrowth and then performs fast pruning of redundant Gaussian points using an opacity-based loss function. After that, it employs a graph-based feature encoding network to extract spatial features from the pruned Gaussian points, guiding the optimization of its spatial distribution. Meanwhile, the incorporation of global alignment loss and local smoothness loss improves spatial consistency and continuity by encouraging Gaussian points with similar features to remain close in Euclidean space. Additionally, other attributes of the Gaussian points (\textit{e.g.}, colors and scale) are jointly updated during spatial optimization.

\begin{figure}[t]
    \centering
    \includegraphics[width=0.9\linewidth]{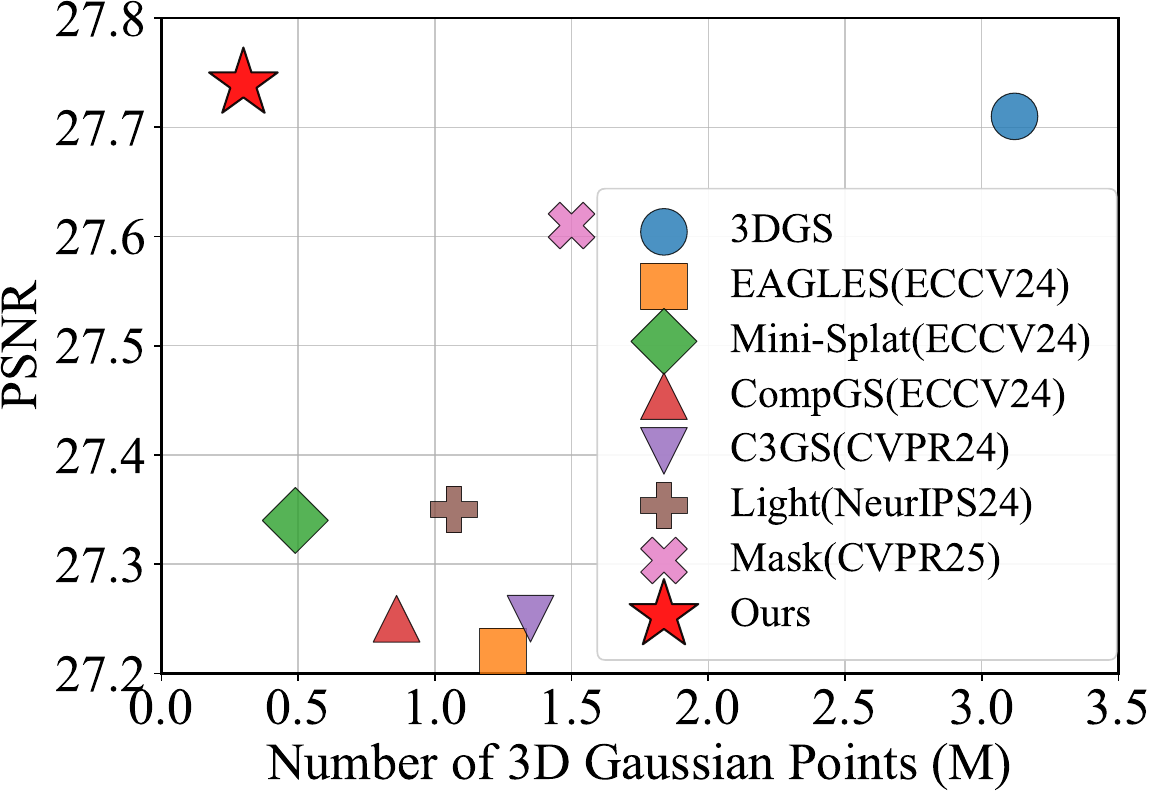}
    \caption{Comparison of rendering quality (PSNR) and \textit{number of Gaussian points} ($N_{\textrm{GS}}$, in millions )  for various methods on the Mip-NeRF360 dataset.}
     \vspace{-4mm}
    \label{fig:mip360}
\end{figure}

For evaluation, we compare the proposed method with the standard 3DGS and state-of-the-art pruning-based variants, on the challenging Mip-NeRF 360~\cite{barron2022mip} and Tanks \&  Temples~\cite{knapitsch2017tanks} datasets. As shown in~\cref {fig:mip360}  and~\cref{table:baseline1},  GS\textasciicircum2   achieves higher PSNR than 3DGS using only about 12.5\% Gaussian points. Moreover, it surpasses all compared baselines in both rendering quality and memory efficiency. 

In summary, our contributions include:
\begin{itemize}
    \item We introduce adaptive densification and opacity-aware pruning to remove  Gaussian points and reduce memory consumption with the minimal render quality loss;

\item we propose a graph-based encoder that leverages feature information to guide the spatial redistribution of Gaussian points, thereby enhancing their spatial consistency and continuity; and

\item we conduct extensive experiments on two challenging benchmark datasets, showing that our method achieves high rendering quality with dramatically reduced number of Gaussian points.
\end{itemize}

\begin{figure*}[htbp]  
    \centering  
    \includegraphics[width=0.9\textwidth]{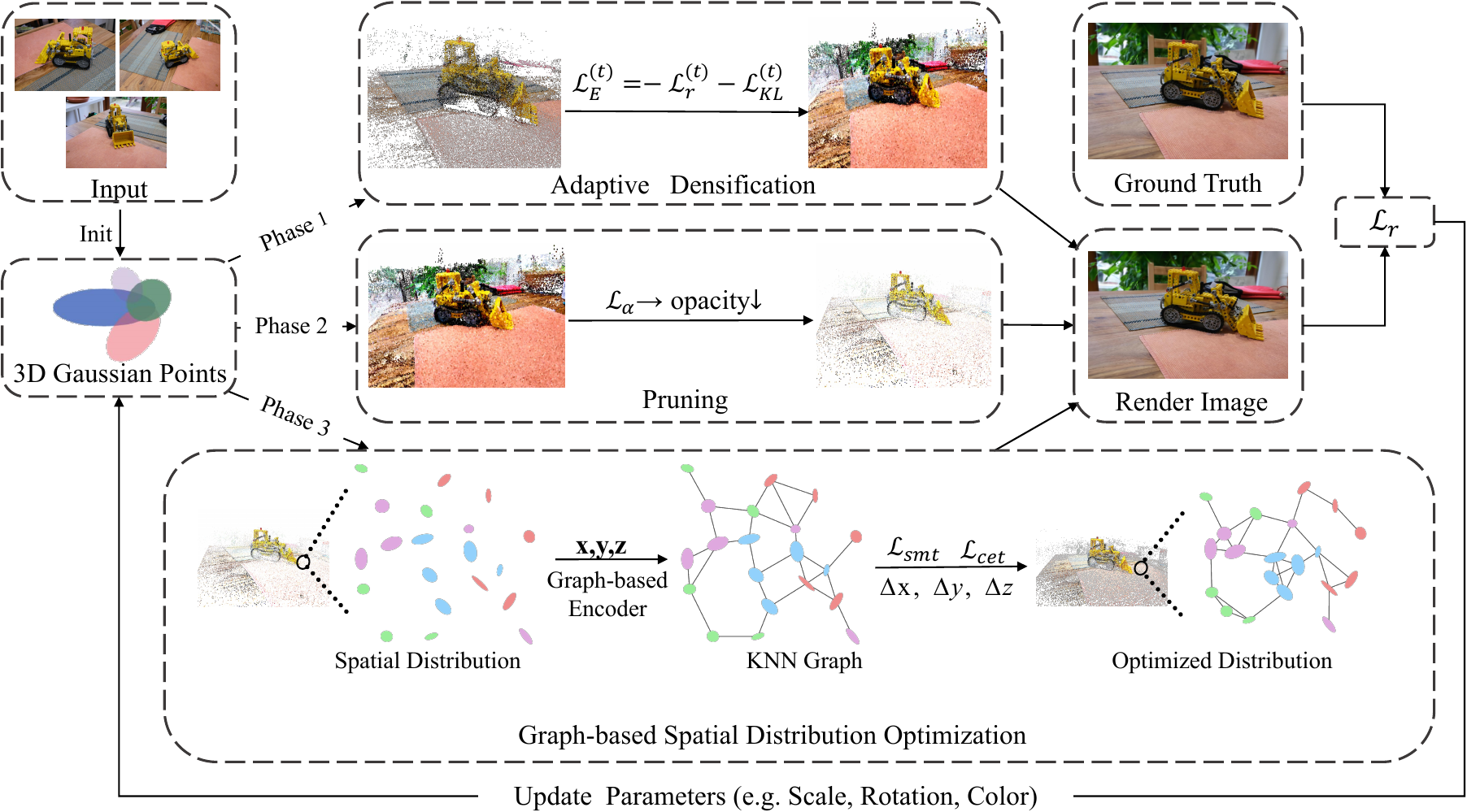}  
    \caption{The pipeline of our method. We propose an \textit{Adaptive Densification and Pruning} (ADP) module that adaptively increases point density in phase 1 and prunes low-opacity points in phase 2.  To improve the spatial distribution after pruning, we introduce a \textit{Graph-based Spatial Distribution Optimization} (GSDO) module, which refines spatial distribution in phase 3 using a lightweight graph-based feature encoder with global alignment and local smoothness losses. 
  }
    \label{fig:over}  
      \vspace{-4mm}
\end{figure*}  

\section{Related Work}
\subsection{Novel View Synthesis}

Novel view synthesis involves generating images from perspectives different from the input views by leveraging existing multi-view images. NeRF~\cite{mildenhall2021NeRF} achieves photorealistic novel view synthesis by leveraging neural networks for implicit scene representation. Despite its impressive rendering performance, NeRF suffers from low computational efficiency due to its reliance on \textit{multilayer perceptrons} (MLPs). Although several works~\cite{insafutdinov2018unsupervised, lin2018learning, yifan2019differentiable, wiles2020synsin, aliev2020neural, kopanas2021point, xu2022point} aim to mitigate these limitations, fundamental issues remain. In contrast, 3DGS \cite{kerbl20233d} has gained significant attention for enabling real-time, high-fidelity rendering. Recent extensions further enhance its applicability, including anti-aliasing techniques \cite{yu2023mip, yan2023multi}, memory optimization~\cite{fan2024lightgaussian, niedermayr2024compressed, navaneet2024compgs,lee2024c3dgs, morgenstern2023compact, lu2024scaffold}, replacement of spherical harmonics for modeling high-frequency reflectance~\cite{yang2024spec}, and dynamic scene modeling~\cite{luiten2023dynamic, yang2023deformable, wu20244d, yang2023real, katsumata2023efficient, kratimenos2023dynmf, huang2023sc}. Although 3DGS offers the aforementioned advantages, it still suffers from high memory overhead.

\subsection{Compact Gaussian Splatting}
3DGS initializes the Gaussian points distribution from a sparse point cloud generated by  SfM and employs an adaptive density control mechanism to prune, split, or clone Gaussian points during optimization. Nevertheless, this mechanism is susceptible to the quality of the initial point cloud and prone to producing redundant Gaussian points. To reduce memory overhead, codebook-based compression \cite{fan2024lightgaussian,lee2024c3dgs,navaneet2024compgs,niedermayr2024compressed} and pruning strategies \cite{fan2024lightgaussian,lee2024c3dgs,niemeyer2024radsplat,zhang2024lp,PUP3DGS} are proposed to quantize Gaussian parameters or discard non-contributing Gaussian points. Anchor-based methods~\cite{lu2024scaffold,chen2024hac} improve spatial efficiency by representing Gaussians with anchors and modeling their relationships to reduce redundancy. Among these approaches, pruning-based 3DGS variants have received particular attention for effectively balancing memory reduction and rendering quality.

Pruning-based methods can be typically divided into two groups: those by combining masking with scale or opacity~\cite{lee2024c3dgs,liu2025maskgaussian}, and those by hand-crafted importance scores~\cite{girish2024eagles,fan2024lightgaussian}.  While these methods achieve competitive compression ratios, they ignore the spatial rearrangement of the remaining Gaussian points after pruning. In practice, the retained Gaussian points are offset and adjusted to compensate for the removed Gaussian points, but the extent to which this implicit rearrangement is able to restore perceptual fidelity remains unexplored. Mini-Splatting~\cite{fang2024mini} improves spatial distribution through densification, effectively reducing memory consumption with minimal quality loss. However, the lack of global and local continuity constraints limits its ability to achieve coherent spatial distributions.

To address this problem, we adopt a graph-based feature encoding network to extract spatial features from pruned Gaussian points and guide their redistribution. Different from previous methods, our solution does not modify the original 3DGS pipeline, such as the representation (anchors). Instead, we focus on minimizing the number of Gaussian points while maintaining high rendering quality.

\section{Method}

\label{sec:method}
 As illustrated in~\cref{fig:over},
we first introduce an \textit{Adaptive Densification and Pruning} (ADP) module to eliminate redundant Gaussian points. However, pruning operations often disrupt spatial continuity and consistency. 
Addressing this issue, we propose a \textit{Graph-Based Spatial Distribution Optimization} (GSDO) module. GSDO represents Gaussian points as a graph to capture spatial and feature relationships, enabling effective redistribution through global alignment and local smoothness constraints. 

\subsection{Preliminaries}
3DGS~\cite{kerbl20233d}  represents a scene by $N$ Gaussian points $\{(\mu_i^{\text{3D}},\Sigma_i,\mathbf{c}_i)\}_{i=1}^N$, where $\mu_i^{\text{3D}}$, $\Sigma_i$ and $\mathbf{c}_i$ denote respectively the 3D center, covariance and color of the $i$-th Gaussian. To enhance rendering quality, 3DGS adopts a block-based strategy and introduces an opacity term $\rho_i(\mathbf{p})$ to better capture real-world appearance, defined as:
\begin{equation}
\rho_i(\mathbf{p}) = \alpha_i G_i,
\end{equation}
where $\alpha_i$ denotes the attribute parameter of the Gaussian point, $\mathbf{p}$ represents its projection onto the pixel, and $G_i$ is the position-dependent attenuation weight. The rendered color $C(\mathbf{p})$ at pixel $\mathbf{p}$ is computed as:
\begin{equation}
    C(\mathbf{p}) = \sum_{i=1}^N \mathbf{c}_i \rho_i(\mathbf{p}) \prod_{j=1}^{i-1} (1 -\rho_i(\mathbf{p})) ,
\end{equation}
where the product term models transmittance, simulating occlusion and light attenuation. The 3DGS loss $\mathcal{L}_r$ combines an $\mathcal{L}1$ loss with a D-SSIM term $\mathcal{L}_{\text{D-SSIM}}$, defined as:
\begin{equation}
 \mathcal{L}_r =  (1 - \lambda_1) \mathcal{L}_1 + \lambda_1 \mathcal{L}_{\text{D-SSIM}}.
 \end{equation}
Specifically, $\mathcal{L}_{\text{D-SSIM}}$ follows the definition in~\cite{wang2004image}, while $\mathcal{L}_1$ is formulated as:
\begin{equation}
\mathcal{L}_1 = \frac{1}{N_{\text{pixel}}} \sum_{i=1}^{N_{\text{pixel}}} \big| \mathbf{I}(x_i, y_i) - \mathbf{I}_{\text{gt}}(x_i, y_i) \big|,
\end{equation}
where $\mathbf{I}(x_i, y_i)$ denotes the color of the rendered image at pixel $(x_i, y_i)$ computed from $C(\mathbf{p})$, $ \mathbf{I}_{\text{gt}}(x_i, y_i)$   represents the corresponding ground-truth color, and $N_{\text{pixel}}$ is the total number of pixels in the image.

\begin{figure}[t]
    \centering
    \includegraphics[width=0.8\linewidth]{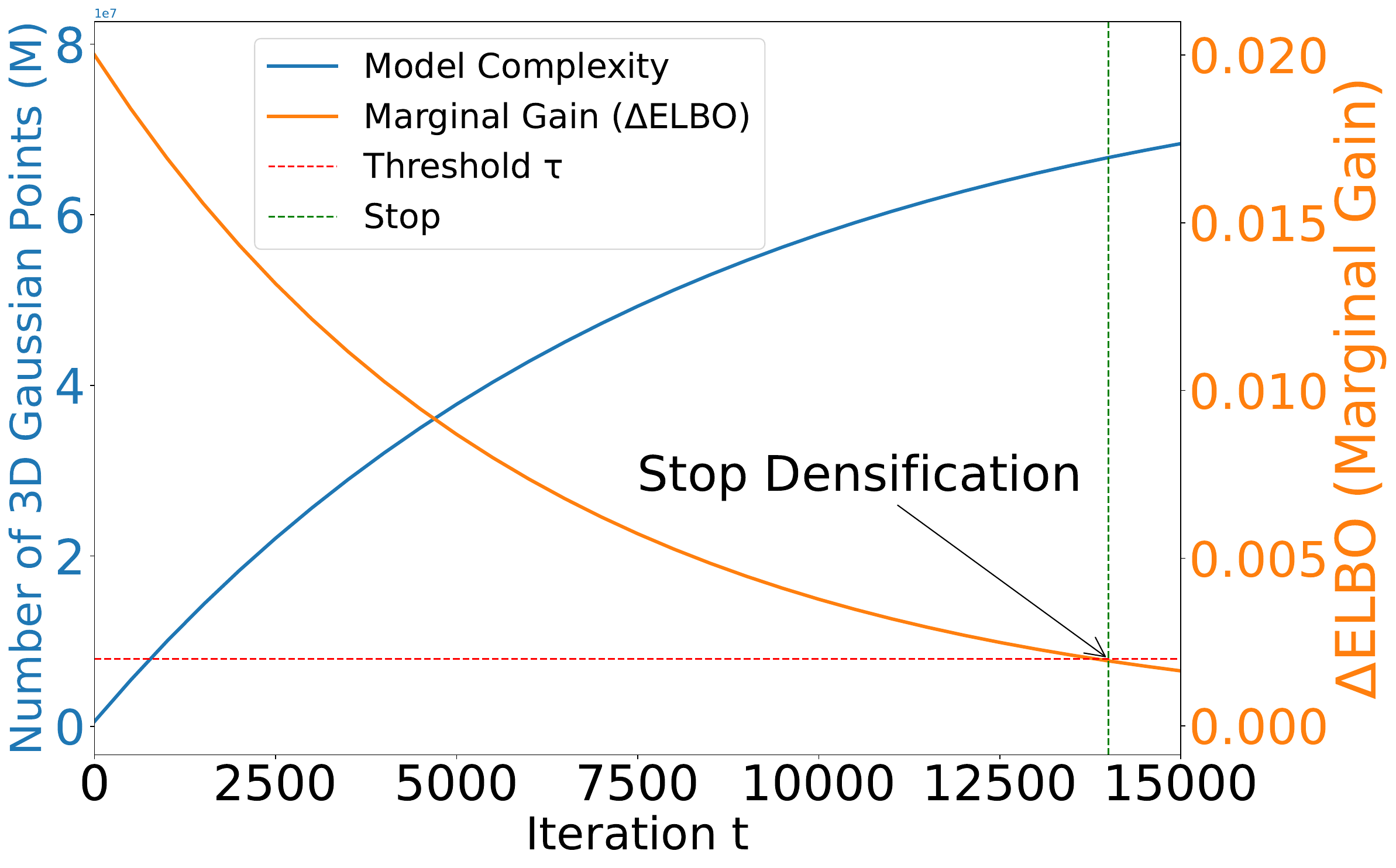}
    \caption{ELBO-guided adaptive densification.}
    \label{fig:ELBO}
      \vspace{-4mm}
\end{figure}

\subsection {Adaptive Densification and Pruning}
\label{subsec: Adaptive Densification and Pruning}

3DGS dynamically adjusts the number of Gaussian points during optimization through densification and pruning. However, existing densification strategies rely on manually defined iteration thresholds. Such strategies lack adaptability and often generate redundant points in simple scenes (\textit{e.g.}, large planar indoor areas), leading to excessive memory overhead. To address this limitation, we introduce an adaptive strategy inspired by the \textit{evidence lower bound} (ELBO)~\cite{kingma2013auto,blei2017variational}.  Our strategy adaptively terminates the 3DGS densification process once the growth in model complexity far surpasses the improvement in rendering quality.  The rendering quality is characterized by the 3DGS loss $\mathcal{L}_r$, formulated as follows:
\begin{equation}
\small
\begin{aligned}
\mathcal{L}_{\mathrm{KL}}^{(t)}
&= \tfrac{1}{2}\!\left[\mathrm{tr}(\tilde{\Sigma}) - \log\lvert \tilde{\Sigma}\rvert\right]
   + \lambda_{\xi}\,\log\!\left(1+\xi\right),\\[3pt]
\mathcal{L}_{\text{E}}^{(t)}
&=
-\mathcal{L}_{\text{r}}^{(t)}
-
\mathcal{L}_{\text{KL}}^{(t)}, \\[3pt]
\end{aligned}
\normalsize
\end{equation}
where $\mathcal{L}_{\text{r}}^{(t)}$  denotes the rendering quality, $\mathcal{L}_{\text{KL}}^{(t)}$ represents model complexity characterized by the 3DGS point density and covariance, and $\mathcal{L}_{\text{E}}^{(t)}$ denotes the ELBO, all at iteration $t$. In addition, $\lambda_{\xi}$ serves as a weighting factor, $\tilde{\Sigma}$ denotes the normalized covariance, and $\xi$ represents the normalized point density. To suppress short-term fluctuations in the ELBO, we apply an \textit{exponential moving average} (EMA) to smooth the ELBO-inspired loss:
\begin{equation}
\small
\begin{aligned}
\widehat{\mathcal{L}}_{\text{E}}^{(t)}
&=
\varepsilon\,\widehat{\mathcal{L}}_{\text{E}}^{(t-1)}
+ (1-\varepsilon)\,\mathcal{L}_{\text{E}}^{(t)}, \\[3pt]
\Delta_t
&=
\frac{
\big|\widehat{\mathcal{L}}_{\text{E}}^{(t)}
- \widehat{\mathcal{L}}_{\text{E}}^{(t-w)}\big|
}{
\big|\widehat{\mathcal{L}}_{\text{E}}^{(t)}\big| 
}, \\[3pt]
\end{aligned}
\normalsize
\end{equation}
where $\varepsilon$ is the smoothing factor, $\widehat{\mathcal{L}}_{\text{E}}^{(t)}$ denotes the smoothed ELBO loss, $\widehat{\mathcal{L}}_{\text{E}}^{(t-1)}$ is the previous value, and $\Delta_t$ represents the relative change of the smoothed ELBO within a $w$-iteration window. When $\Delta_t$ converges to $\tau$, it indicates that further improvements in rendering quality are negligible compared to the increase in model complexity. Consequently, the 3DGS densification process is automatically terminated, as illustrated in~\cref{fig:ELBO}.
\begin{figure}
    
    \centering
    \includegraphics[width=0.7\linewidth]{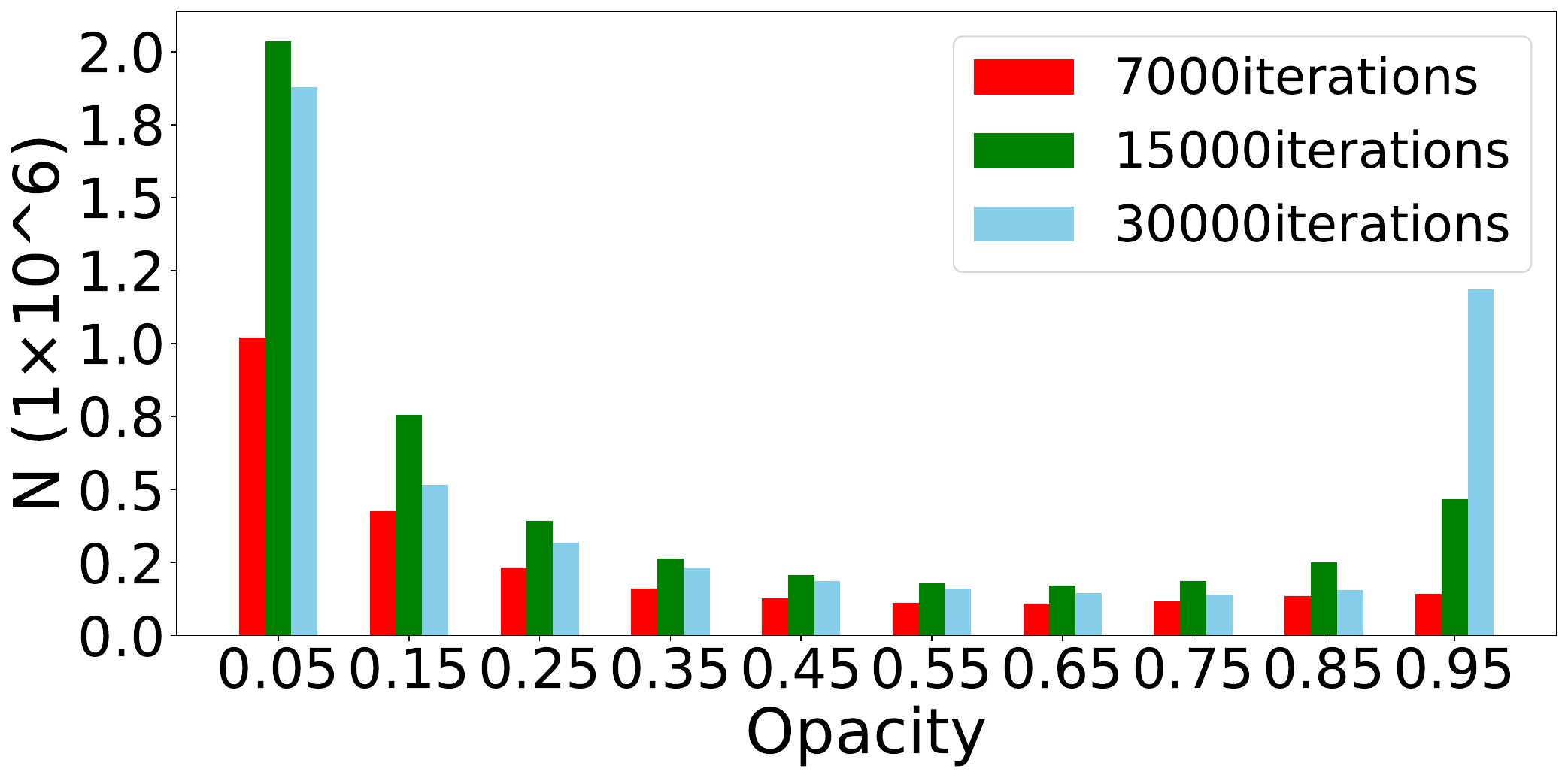}
\caption{Analyzing the impact of optimization.
Low-opacity Gaussian points ($\text{opacity} \leq 0.1$) constitute a considerable portion of the total. For instance, in the Lighthouse scene from the Tanks~\&~Temples dataset, they account for approximately 40\% of all Gaussian points.}
    \label{fig:enter-label}
      \vspace{-4mm}
\end{figure}

\begin{figure*}[htbp]
    \centering
    \small
    \begin{tabular}{@{}c@{\hspace{0.5mm}}c@{\hspace{1mm}}c@{\hspace{1mm}}c@{\hspace{1mm}}c@{}}
        \raisebox{0.1\height}{\rotatebox{90}{\textbf{Render View}}} &
        \includegraphics[width=0.23\textwidth]{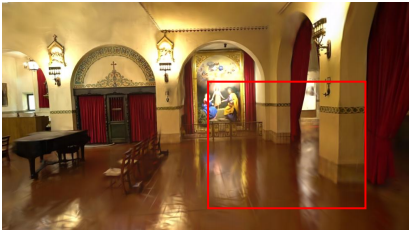} &
        \includegraphics[width=0.23\textwidth]{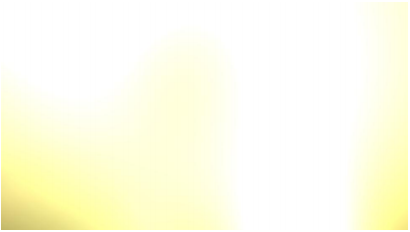} &
        \includegraphics[width=0.23\textwidth]{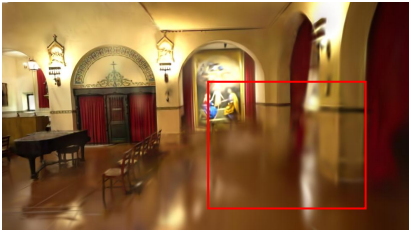} &
        \includegraphics[width=0.23\textwidth]{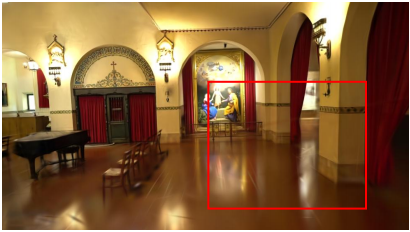} \\[1pt]

        \raisebox{0.45\height}{\rotatebox{90}{\textbf{GS View}}} &
        \includegraphics[width=0.23\textwidth]{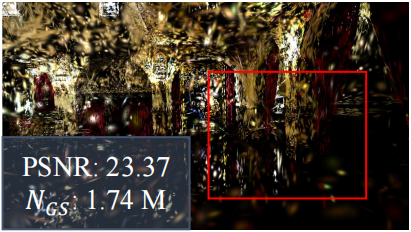} &
        \includegraphics[width=0.23\textwidth]{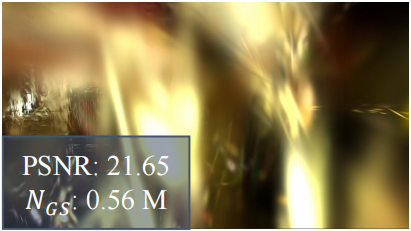} &
        \includegraphics[width=0.23\textwidth]{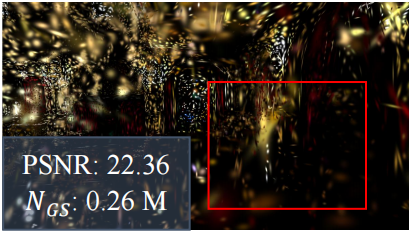} &
        \includegraphics[width=0.23\textwidth]{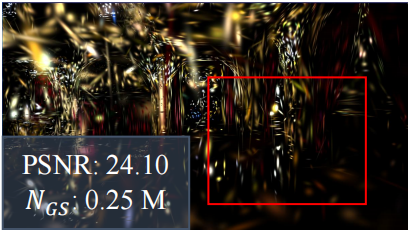} \\[-2pt]

        & \small (a) 3DGS & \small (b) LightGaussian & \small (c) Mini-Splatting & \small (d) Ours
    \end{tabular}
    \caption{Impact of pruning. 
    We report PSNR and $N_{\textrm{GS}}$ (in millions) for each method on the church scene from Tanks~\&~Temples. Pruning degrades the spatial consistency and continuity of the Gaussian points, leading to visible artifacts.}
    \label{fig:analysis}
      \vspace{-4mm}
\end{figure*}

Although the ELBO-guided strategy effectively reduces memory consumption, many low-opacity Gaussian points still persist, contributing little to rendering quality while occupying substantial memory, as shown in~\cref{fig:enter-label}. For this problem, we introduce an opacity-aware regularization loss:
\begin{equation}
\mathcal{L}_\alpha = \sum_{i=1}^{N} \left( \lambda_2 \rho_i^2 + \lambda_3 \rho_i \right),
\end{equation}
where $\lambda_2$ and $\lambda_3$ are regularization weights. The linear term encourages faster convergence of low-opacity Gaussian points. In our experiments, we observed that certain blur artifacts arise from anomalous Gaussian points with excessively high opacity, which cannot be effectively suppressed by the linear term alone or by simply increasing the pruning threshold. To address this issue, we introduce a higher-order term that imposes a stronger penalty on such cases. 

During 3DGS training, we employ a dynamic opacity-based pruning strategy. Every 100 iterations, Gaussian points with opacity below a predefined threshold are pruned. Compared to LightGaussian, our approach removes more redundant Gaussian points and effectively suppresses artifacts caused by anomalous high-opacity points, as shown in~\cref{table:baseline1} and~\cref{fig:analysis}.


\subsection{Spatial Distribution Analysis of 3DGS}
\label{subsec: Analyzing 3DGS From Spatial Distribution}
The pruning strategy removes a large number of low-impact Gaussian points while effectively preserving the overall structural integrity of both objects and backgrounds. However, some local regions still suffer from insufficient point coverage, leading to noticeable deviations from the ground truth. As shown in~\cref{table:ablation1}, pruning causes a substantial drop in rendering quality, with a PSNR decrease of up to 0.94 on the Mip-NeRF360 dataset.

The primary reason is that pruning removes a large number of Gaussian points, causing the remaining ones to migrate toward uncovered regions to compensate for the missing points. However, this process disrupts spatial continuity, making the movement directions of the remaining Gaussian points potentially inaccurate. Moreover, not all points should shift toward underrepresented areas—some should remain stationary to preserve local geometric fidelity. As shown in~\cref{fig:teaser} and~\cref{fig:analysis}, the loss of spatial continuity caused by pruning prevents 3DGS-based methods from effectively optimizing the positions of the remaining points, leading to noticeable blurring artifacts. Furthermore, as presented in~\cref{table:ablation1}, simply increasing the number of training iterations yields limited improvement in spatial arrangement. These findings indicate that the original 3DGS optimization process is insufficient to restore spatial consistency after pruning. Thus motivated, we improve rendering quality by explicitly optimizing the spatial distribution of 3DGS.

\subsection{Graph-based Spatial Distribution Optimization}
\label{subsec: Spatial distribution optimization}

During the pruning process in 3DGS, removing a large number of Gaussian points often results in sparse and uneven point distributions, thereby disrupting spatial consistency. To enhance the spatial continuity of the remaining Gaussian points, we introduce a lightweight graph-based encoder inspired by 4DGS~\cite{wu20244d}. While 4DGS focuses on modeling spatial–temporal continuity for dynamic scenes, its high computational complexity and spatial sensitivity limit its applicability to static settings. In contrast, our lightweight graph-based encoder extracts structural features from the pruned Gaussian points to guide their spatial redistribution during optimization. Unlike 4DGS, our approach does not rely on spatial–temporal consistency and maintains low computational overhead.

\begin{table*}[!t]
\small
\centering
\begin{tabular}{
  @{\hspace{0.5mm}}c@{\hspace{1mm}}   
  l@{\hspace{2mm}}                    
  l@{\hspace{0.5mm}}l@{\hspace{0.5mm}}l@{\hspace{0.5mm}}
  l@{\hspace{3mm}}               
  l@{\hspace{0.5mm}}
  l@{\hspace{0.5mm}}
  l@{\hspace{0.5mm}}
  l@{\hspace{3mm}}                   
  l@{\hspace{0.5mm}}
  l@{\hspace{0.5mm}}
  l@{\hspace{0.5mm}}
  l
}
 \hline
   \multirow{2}{*}{Dataset } & \multirow{2}{*}{Method }
     & \multicolumn{4}{c}{All Scenes}
     & \multicolumn{4}{c}{Outdoor}
     & \multicolumn{4}{c}{Indoor} \\
   \cline{3-14}
    & & PSNR$\uparrow$ & SSIM$\uparrow$ & LPIPS$\downarrow$ & $N_{\textrm{GS}}$$\downarrow$
      & PSNR$\uparrow$ & SSIM$\uparrow$ & LPIPS$\downarrow$ & $N_{\textrm{GS}}$$\downarrow$
      & PSNR$\uparrow$ & SSIM$\uparrow$ & LPIPS$\downarrow$ & $N_{\textrm{GS}}$$\downarrow$ \\
  \hline
  & 3DGS & \underline{27.71} & \textbf{0.83} & \textbf{0.20} & 3.12
        & \underline{24.99} & \textbf{0.74} & \textbf{0.23} & 4.50
        & \underline{31.09} & \underline{0.92} & \underline{0.19} & 1.41 \\
  \multirow{5}{*}{\rotatebox[origin=c]{90}{Mip-NeRF 360}}
    & EAGLES         & 27.22 & \underline{0.82} & 0.23 & 1.25
      & 24.64 & \underline{0.73} & 0.27 & 1.69
      & 30.46 & \textbf{0.93} & \textbf{0.18} & 0.71 \\
    & Mini-Splatting & 27.34 & \underline{0.82} & \underline{0.22} & \underline{0.49}
      & 24.71 & \textbf{0.74} & \underline{0.24} & \underline{0.57}
      & 30.61 & \underline{0.92} & \underline{0.19} & \underline{0.40} \\
    & CompGS         & 27.25 & \underline{0.82} & 0.23 & 0.86
      & 24.79 & \underline{0.73} & 0.26 & 1.19
      & 30.33 & \underline{0.92} & \underline{0.19} & 0.46 \\
    & Compact 3DGS  & 27.25 & 0.81            & 0.23 & 1.35
      & 24.61 & 0.72            & 0.27 & 1.47
      & 30.54 & \underline{0.92} & \textbf{0.18} & 1.21 \\
    & LightGaussian  & 27.35 & \underline{0.82} & 0.23 & 1.07
      & 24.90 & \underline{0.73} & 0.26 & 1.17
      & 30.41 & \underline{0.92} & \underline{0.19} & 0.94 \\
    & MaskGaussian   & 27.61 & \underline{0.82} & 0.23 & 1.50
      & 24.98 & \textbf{0.74}    & 0.26 & 2.29
      & 30.91 & \textbf{0.93}    & \underline{0.19} & 0.51 \\
    & Ours           & \textbf{27.74} & \underline{0.82} & 0.23 & \textbf{0.30}
      & \textbf{25.01} & \underline{0.73} & 0.27 & \textbf{0.32}
      & \textbf{31.16} & \textbf{0.93} & \textbf{0.18} & \textbf{0.27} \\
  \hline
  & 3DGS           & \underline{24.19} & \underline{0.84} & \textbf{0.19} & 1.57
      & \underline{23.68} & \textbf{0.84} & \textbf{0.21} & 1.27
      & \underline{25.00} & \textbf{0.86} & \textbf{0.17} & 2.06 \\
  \multirow{5}{*}{\rotatebox[origin=c]{90}{T\&T}}
    & EAGLES         & 23.78 & \underline{0.84} & 0.21 & 0.61
      & 23.30 & \underline{0.83} & \underline{0.22} & 0.53
      & 24.56 & \underline{0.85} & 0.19 & 0.75 \\
    & Mini-Splatting & 23.72 & \underline{0.84} & 0.21 & \underline{0.27}
      & 23.30 & \underline{0.83} & \underline{0.22} & \underline{0.26}
      & 24.40 & \underline{0.85} & 0.19 &\textbf{0.27} \\
    & CompGS         & 23.84 & \underline{0.84} & 0.21 & 0.55
      & 23.32 & \underline{0.83} & 0.23 & 0.48
      & 24.68 & \underline{0.85} & 0.19 & 0.66 \\
    & Compact 3DGS  & 23.76 & 0.83            & 0.22 & 0.80
      & 23.42 & 0.82            & 0.23 & 0.64
      & 24.32 & 0.84            & 0.20 & 1.06 \\
    & LightGaussian  & 23.16 & 0.82            & 0.25 & 0.53
      & 22.87 & 0.81            & 0.26 & 0.47
      & 23.65 & 0.83            & 0.23 &\underline{0.63} \\
    & MaskGaussian   & 24.16 & \underline{0.84} & \underline{0.20} & 0.74
      & 23.65 & \underline{0.83} & \textbf{0.21} & 0.61
      & 24.98 & \textbf{0.86}    & \underline{0.18} & 0.96 \\
    & Ours           & \textbf{24.90} & \textbf{0.85} & 0.21 & \textbf{0.24}
      & \textbf{24.58} & \textbf{0.84} & \underline{0.22} & \textbf{0.22}
      & \textbf{25.42} & \textbf{0.86} & \underline{0.18} & \textbf{0.27} \\
  \hline
\end{tabular}
\caption{Quantitative rendering results of Mip-NeRF 360 and T\&T. \textbf{Bold} font highlights the best results and \underline{underline} for the second.}
\label{table:baseline1}
  \vspace{-5mm}
\end{table*}

Specifically, our encoder maps each Gaussian point coordinate $\mathbf{x}_i \in \mathbb{R}^3$ to a latent feature vector $\mathbf{z}_i \in \mathbb{R}^D$, which provides guidance for subsequent spatial optimization in 3DGS. We first compute an initial embedding for each Gaussian center using a single-layer perceptron:
\begin{equation}
\mathbf{f}_i = \sigma(\mathbf{W}_1 \mathbf{x}_i + \mathbf{b}_1),
\end{equation}
where $\mathbf{W}_1 \in \mathbb{R}^{D \times 3}$ and $\mathbf{b}_1 \in \mathbb{R}^D$ are learnable parameters, and $\sigma(\cdot)$ denotes a nonlinear activation function. The resulting feature vector $\mathbf{f}_i \in \mathbb{R}^D$ encodes the initial local geometry of the $i$-th Gaussian center. To capture local geometric relationships, we construct a $k$-nearest neighbor graph $\mathcal{N}_k(i)$ for each point based on the initial embeddings $\mathbf{f}$. We then compute the pairwise residuals as:
\begin{equation}
\Delta_{ij} = \mathbf{f}_i - \mathbf{f}_j,
\end{equation}
and aggregate them as:
\begin{equation}
\mathbf{r}_i = \sum_{j \in \mathcal{N}_k(i)} \Delta_{ij},
\end{equation}
where $\mathbf{r}_i \in \mathbb{R}^D$ captures the local geometric offset around point $i$. It is then refined through a second transformation:
\begin{equation}
\mathbf{h}_i = \text{ReLU}\left( \mathbf{W}_2 (\mathbf{f}_i + \mathbf{r}_i) + \mathbf{b}_2 \right),
\end{equation}
where $\mathbf{W}_2 \in \mathbb{R}^{D \times D}$ and $\mathbf{b}_2 \in \mathbb{R}^D$ are additional learnable parameters, and $\mathbf{h}_i$ denotes the residual-enhanced local feature. To integrate global context, we further apply local max pooling over the $k$-nearest neighbors of each point:
\begin{equation}
\mathbf{m}_i = \max_{j \in \mathcal{N}_k(i)} \mathbf{f}_j,
\end{equation}
and then compute the scene-level global feature as the average of all local max-pooled features:
\begin{equation}
\bar{\mathbf{m}} = \frac{1}{N_{\textrm{GS}}} \sum_{i=1}^{N_{\textrm{GS}}} \mathbf{m}_i,
\end{equation}
where $N_{\textrm{GS}}$ denotes the total number of Gaussian points after pruning. We concatenate $\mathbf{h}_i$ with $\bar{\mathbf{m}}$ and feed them into two fully connected layers to obtain the latent feature vector $\mathbf{z}_i$. By combining local residual aggregation with global contextual information, each Gaussian point center is updated not only based on its local neighborhood but also with respect to the overall scene structure.

Inspired by the spatial constraints used in the 3D scene segmentation method Gaussian Grouping~\cite{10.1007/978-3-031-73397-0_10}, we design two unsupervised loss functions to enhance the spatial consistency of 3DGS. While Gaussian Grouping achieves semantic aggregation through local feature-space clustering, its loss formulation neglects the global spatial distribution in Euclidean space, limiting its ability to maintain overall spatial continuity. In contrast, our loss design jointly enforces global and local constraints, enabling more holistic spatial consistency among the pruned Gaussian points. We compute $\mathcal{L}_{\text{cet}}$ in $\mathbb{R}^3$ via a lightweight projection 
$g:\mathbb{R}^D\!\rightarrow\!\mathbb{R}^3$, i.e., $\hat{x}_i=g(\mathbf{z}_i)$, 
and align the centroids as follows:
\begin{equation}
\mathcal{L}_{\text{cet}} = \Big\| \frac{1}{N_{\textrm{GS}}} \sum_{i=1}^{N_{\textrm{GS}}} \big( \mathbf{x}_i - \hat{x}_i \big) \Big\|_2^2.
\end{equation}
This loss encourages global alignment between the centroids of the Gaussian points in Euclidean and feature spaces. To further enhance local spatial coherence, we introduce a local smoothness loss:
\begin{equation}
\mathcal{L}_{\text{smt}} = \! \frac{1}{M (K-1)} \!\! \sum_{m=1}^{M} \!\! \sum_{i=1}^{K-1} \!\!\big\| \mathbf{z}_i^{(m)} - \mathbf{z}_{i+1}^{(m)} \big\|_2  \big\| \mathbf{x}_i^{(m)} - \mathbf{x}_{i+1}^{(m)} \big\|_2,
\end{equation}
where $M$ denotes the number of local neighborhoods sampled across the 3DGS, and $K$ is the number of Gaussian points within each neighborhood. $\mathbf{x}_i^{(m)}$ and $\mathbf{z}_i^{(m)}$ represent the geometric and feature embeddings of the $i$-th point in the $m$-th neighborhood, respectively. This loss penalizes discrepancies between neighboring points in both Euclidean and feature spaces, effectively suppressing spatial discontinuities and promoting local smoothness.

The final spatial consistency loss combines both terms with an additional regularization component:
\begin{equation}
\mathcal{L}_{\text{final}} = \lambda_c \mathcal{L}_{\text{cet}} + \lambda_s \mathcal{L}_{\text{smt}} + \mathcal{L}_{r},
\end{equation}
where $\lambda_c$ and $\lambda_s$ are regularization weights. This composite loss effectively enforces both global alignment and local smoothness, ensuring a more consistent spatial distribution of Gaussian points after pruning.

\begin{figure*}[h!]
    \centering
    \small
    \setlength{\tabcolsep}{4pt}
    \renewcommand{\arraystretch}{0.5}
    {
    \begin{tabular}
    {@{\hspace{.1mm}}c@{\hspace{.1mm} } @{\hspace{.1mm}}c@{\hspace{.1mm} } @{\hspace{.1mm}}c@{\hspace{.1mm} } @{\hspace{.1mm}}c@{\hspace{.1mm} }}
        \includegraphics[width=0.245\linewidth]{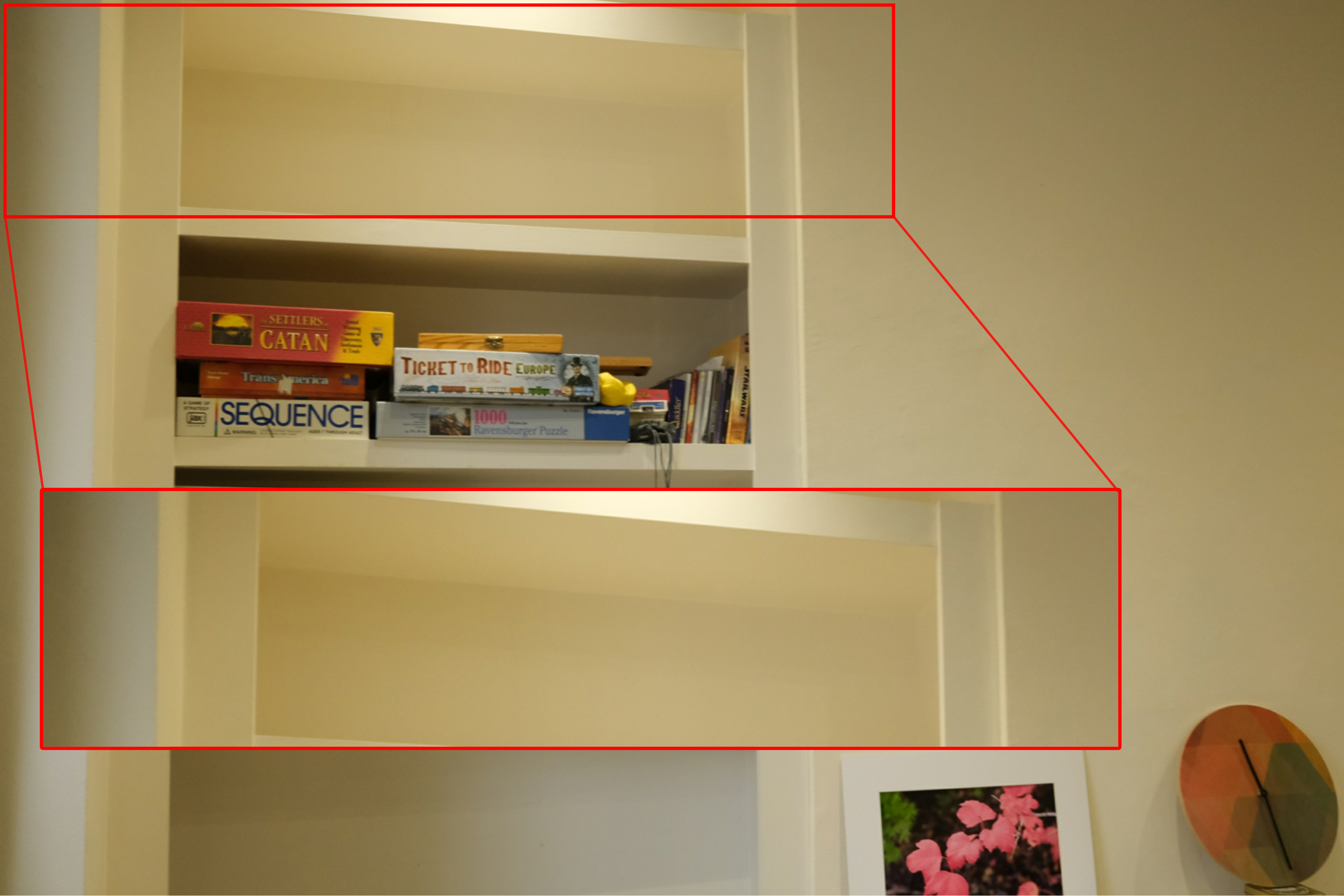}       &
        \includegraphics[width=0.245\linewidth]{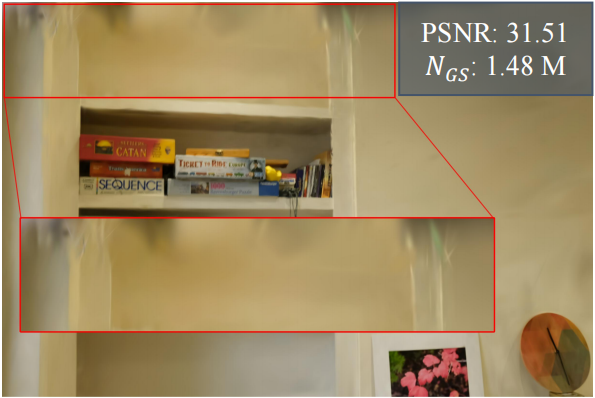}     &
        \includegraphics[width=0.245\linewidth]{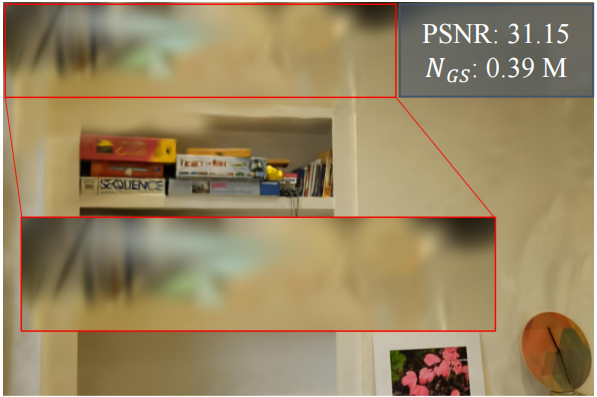}     &
        \includegraphics[width=0.245\linewidth]{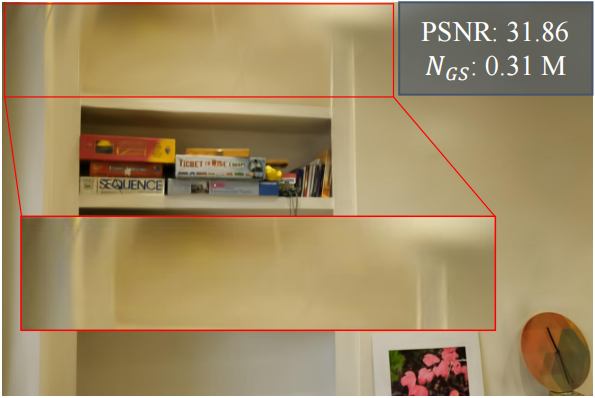}     \\

        \includegraphics[width=0.245\linewidth]{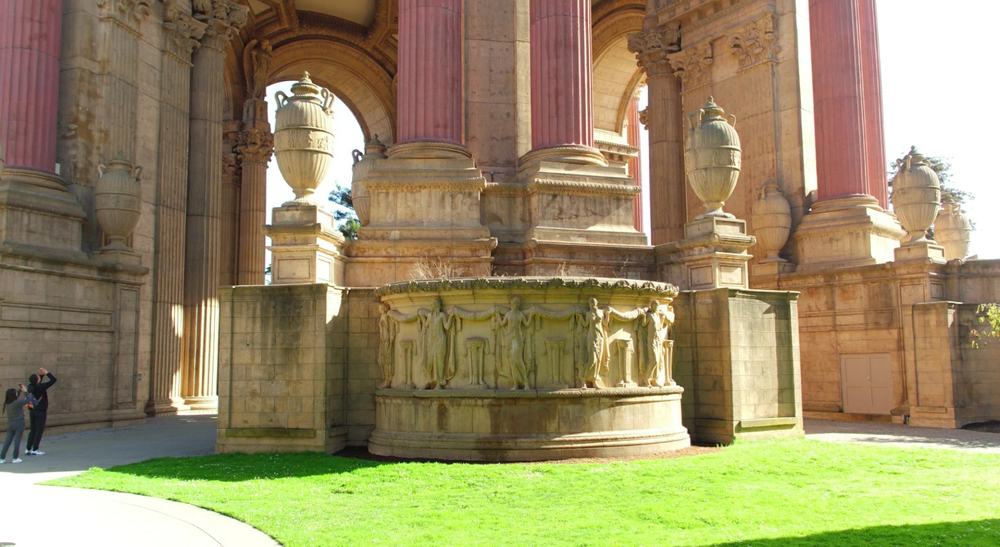}       &
        \includegraphics[width=0.245\linewidth]{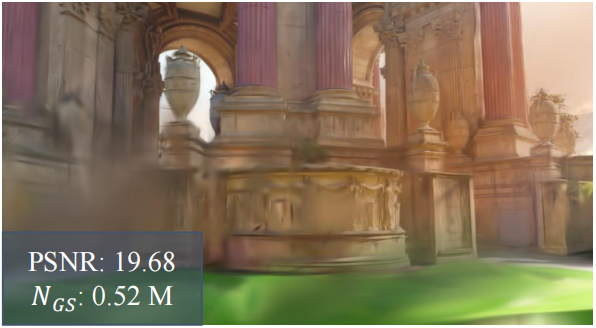}     &
        \includegraphics[width=0.245\linewidth]{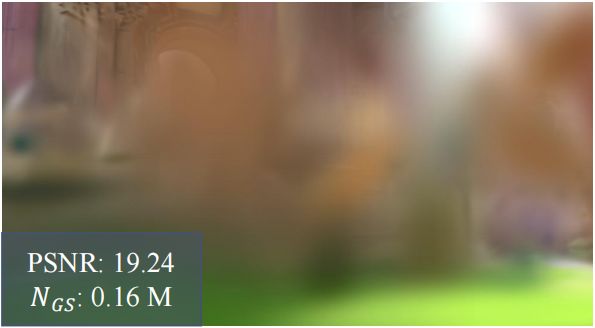}     &
        \includegraphics[width=0.245\linewidth]{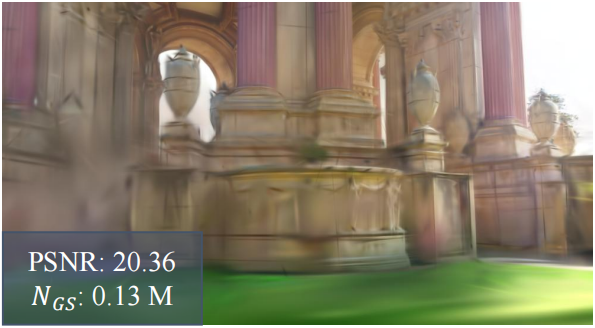}     \\

          \includegraphics[width=0.245\linewidth]{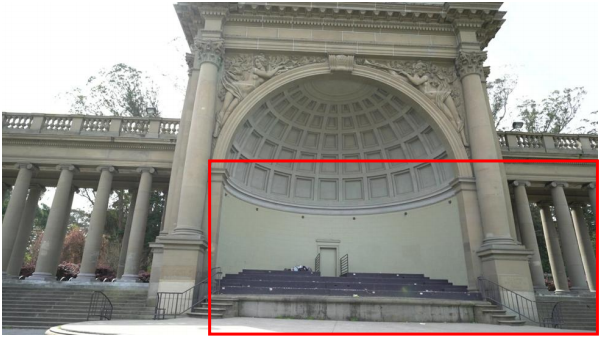}       &
              \includegraphics[width=0.245\linewidth]{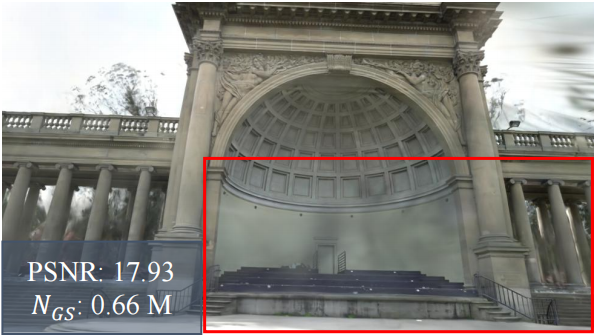}     &
        \includegraphics[width=0.245\linewidth]{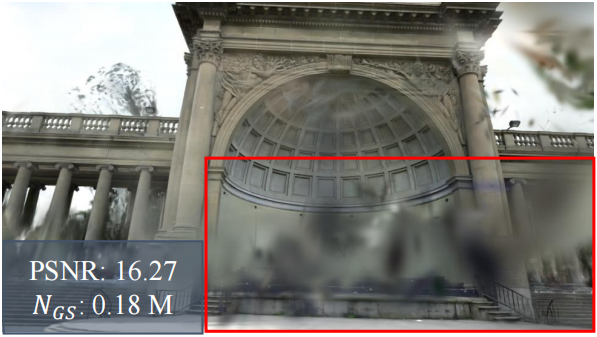}     &
        \includegraphics[width=0.245\linewidth]{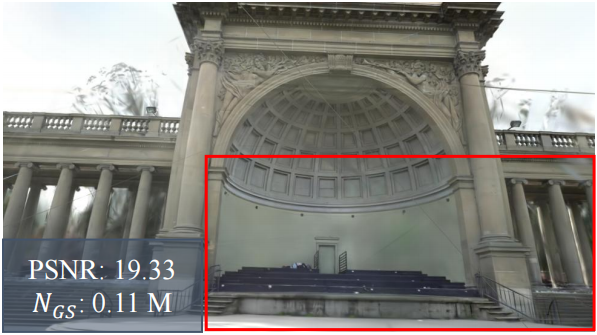}     \\
        
     \includegraphics[width=0.245\linewidth]{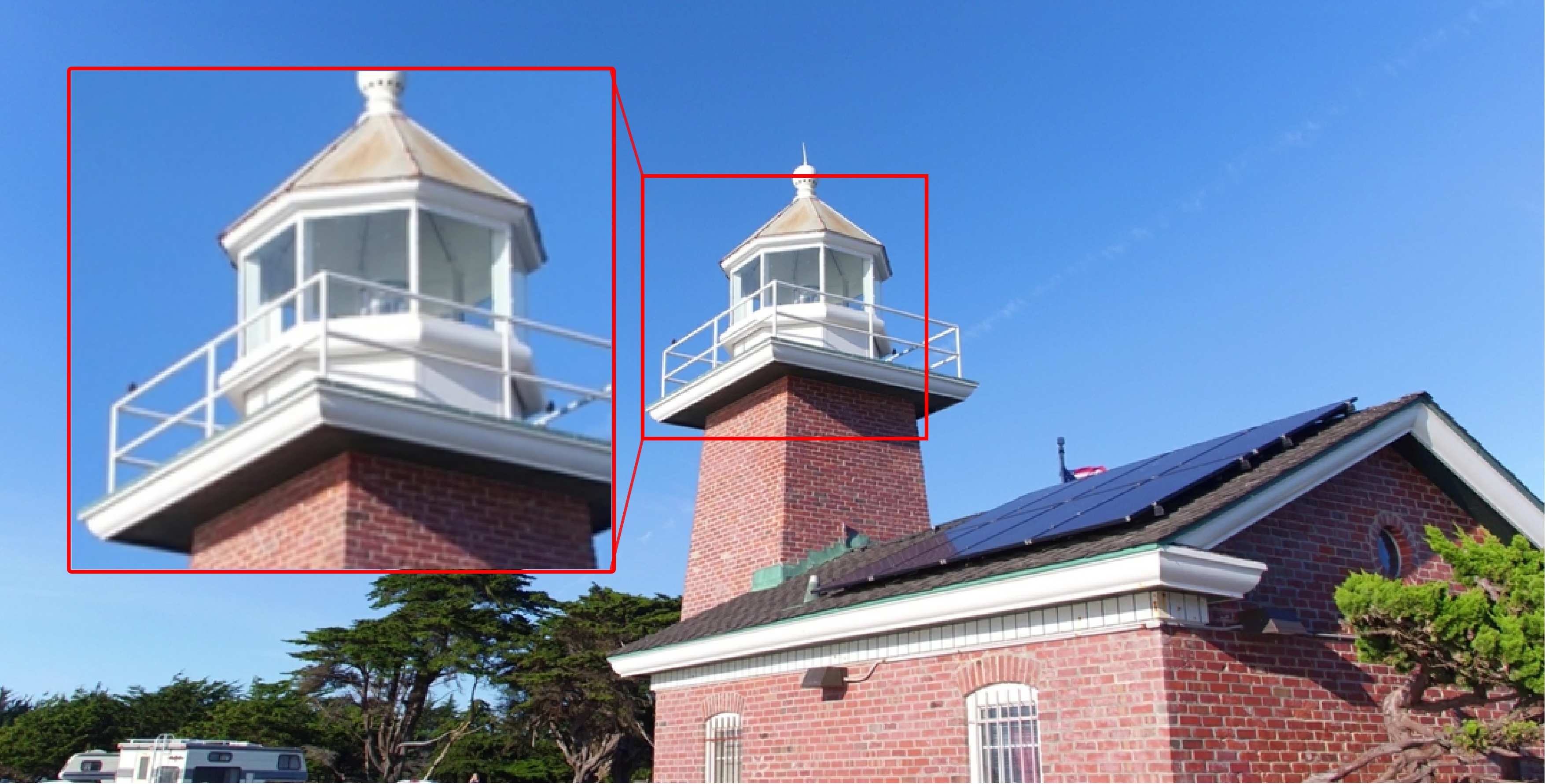}       &
                \includegraphics[width=0.245\linewidth]{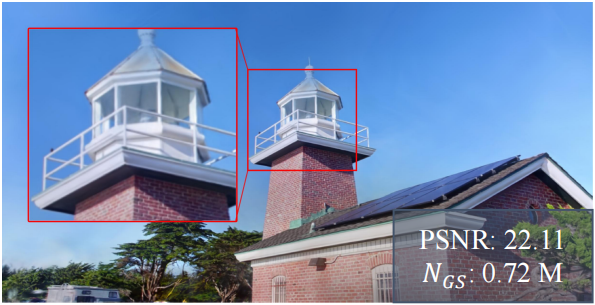}     &
        \includegraphics[width=0.245\linewidth]{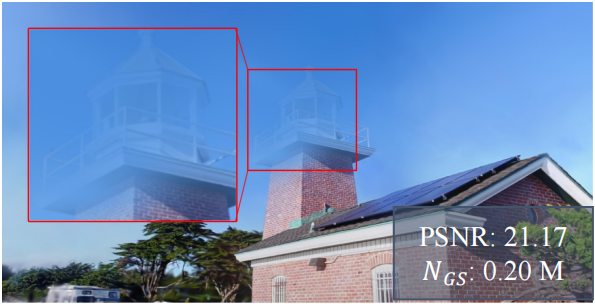}     &
        \includegraphics[width=0.245\linewidth]{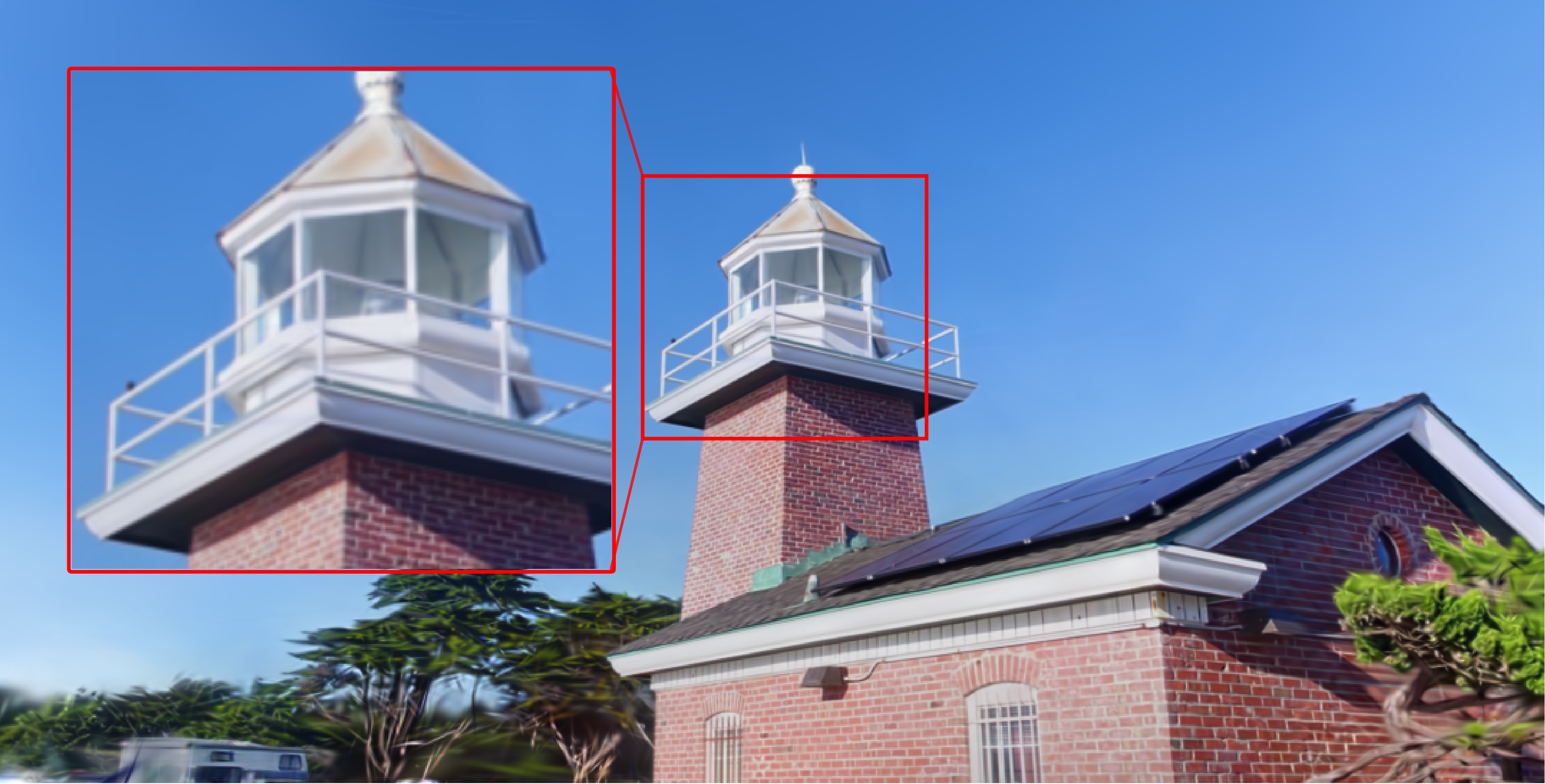}     \\

     \includegraphics[width=0.245\linewidth]{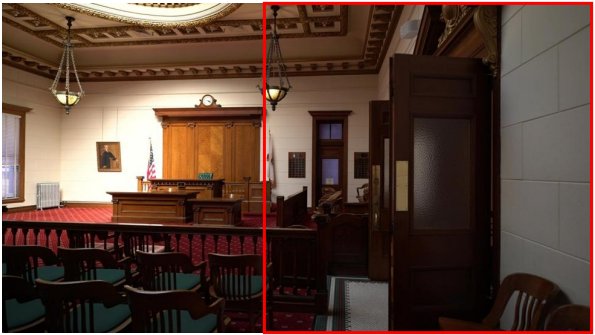}       &
              \includegraphics[width=0.245\linewidth]{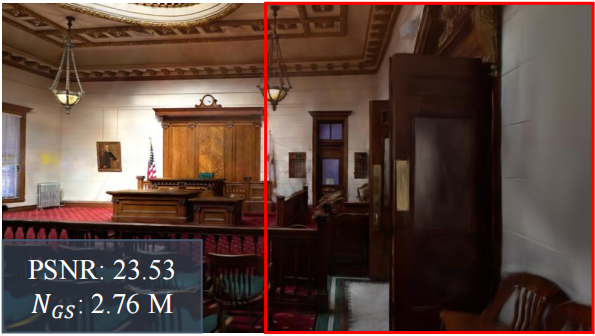}     &
        \includegraphics[width=0.245\linewidth]{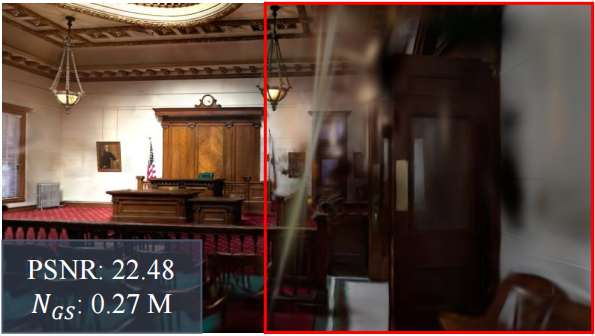}     &
        \includegraphics[width=0.245\linewidth]{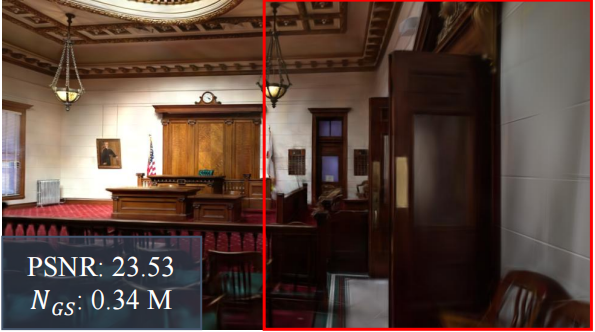}     \\

        (a) GT & (b) 3DGS & (c) Mini-Splatting & (d) Ours \\
    \end{tabular}}
    \vspace{-1mm}
    \captionof{figure}{
      Qualitative comparison. The first scene (\textit{Room}) is from the Mip-NeRF 360 dataset, followed by four scenes (\textit{Palace, Temple, Lighthouse, Courtroom }) from the Tanks~\&~Temples dataset. Key regions have been zoomed in to highlight visual differences. We report the PSNR and $N_{\textrm{GS}}$ (in millions) for each method. 
    }
    \label{fig:baseline1}
    \vspace{-2mm}
\end{figure*}

\section{Experiments}

\subsection{Experimental Settings}
\noindent\textbf{Dataset and Metrics.} We evaluate our method on 30 real-world scenes from publicly available datasets, including all 9 scenes from Mip-NeRF 360~\cite{barron2022mip} and 21 scenes from Tanks~\&~Temples~\cite{knapitsch2017tanks}. These scenes are grouped into indoor (12 scenes) and outdoor scenes (18 scenes). We evaluate performance using four standard metrics: \textit{peak signal-to-noise ratio} (PSNR$\uparrow$), \textit{structural similarity index} (SSIM$\uparrow$)~\cite{wang2004image}, \textit{learned perceptual block similarity} (LPIPS$\downarrow$)~\cite{zhang2018perceptual} and the \textit{number of Gaussian points} ($N_{\textrm{GS}}$$\downarrow$).

\vspace{1mm}\noindent\textbf{Compared Baselines.}
In addition to the original 3DGS~\cite{kerbl20233d}, the present study employs a benchmarking approach against two hand-crafted importance pruning methods, EAGLES~\cite{girish2024eagles} and LightGaussian~\cite{fan2024lightgaussian}, as well as two learnable-mask-based competitors, Compact3DGS~\cite{lee2024c3dgs} and MaskGaussian~\cite{liu2025maskgaussian}. Besides, we also add Mini-Splatting~\cite{fang2024mini} and CompGS~\cite{navaneet2024compgs}, both of which aim to minimize the number of Gaussian points. For fair comparison, our evaluation focuses solely on the reduction of Gaussian points, excluding additional components such as SFM initialization, attribute-vector quantization and spherical-harmonics distillation.

\subsection{Comparison with Baselines}
\label{subsec: Comparison with Baselines}

We provide comprehensive quantitative comparisons in~\cref{table:baseline1}. 
On Mip-NeRF360~\cite{barron2022mip} and T\&T~\cite{knapitsch2017tanks}, GS\textasciicircum2 achieves higher PSNR than 3DGS while using only 9.62\% and 15.29\% of the Gaussian points, respectively.

We attribute this improvement to the pruning process, which effectively mitigates overfitting. Moreover, GS\textasciicircum2 consistently outperforms all baselines in terms of PSNR while using fewer Gaussian points. The performance gain is particularly notable on the T\&T dataset, owing to its larger proportion of indoor scenes, where the geometric and spatial constraints of bounded environments facilitate more effective spatial distribution optimization.

\noindent\textbf{Qualitative Results.}~\cref{fig:analysis} and~\cref{fig:baseline1} present qualitative comparisons among 3DGS~\cite{kerbl20233d}, Mini-Splatting~\cite{fang2024mini} and our proposed method. We evaluate six representative scenes, including three indoor scenes (Church, Room and Courtroom) and three outdoor scenes (Palace, Temple and Lighthouse), to comprehensively assess generalization across diverse spatial structures and lighting conditions. Compared with the original 3DGS, our approach substantially reduces the number of Gaussian points required for high-quality reconstruction. While Mini-Splatting improves spatial distribution through densification, its lack of global and local continuity constraints limits spatial coherence. In contrast, GS\textasciicircum2 not only enhances rendering quality with a compact Gaussian representation but also preserves fine-grained scene structures, achieving robust performance in both indoor and outdoor settings.

\subsection{Ablation Studies}
\label{subsec: Ablation Studies}
To evaluate the contribution of each component, we conduct a comprehensive ablation study on both datasets, as summarized in~\cref{table:ablation1}. Our ADP module comprises two key components: ELBO-based adaptive densification and opacity-based pruning, which together produce a more compact 3DGS representation. Increasing the number of training iterations yields only marginal improvements, particularly in SSIM and LPIPS. To further enhance reconstruction quality, we introduce $\mathcal{L}_{\text{smt}}$ and $\mathcal{L}_{\text{cet}}$, which guide the spatial distribution optimization of 3DGS using feature information to improve spatial consistency and continuity. These results validate the effectiveness of each component and underscore the importance of their joint contribution.

\begin{table}[!t]
\small
\centering
\begin{tabular}{@{\hspace{0.5mm}}l@{\hspace{0.5mm}}l@{\hspace{0.5mm}}l@{\hspace{0.5mm}}l@{\hspace{0.5mm}}l@{\hspace{0.5mm}}l@{\hspace{0.5mm}}}
    \hline
    Dataset & Method & PSNR$\uparrow$ & SSIM$\uparrow$ & LPIPS$\downarrow$ & $N_{\textrm{GS}}$$\downarrow$ \\
    \hline
    \multirow{6}{*}{Mip-NeRF 360}
    & 3DGS                          & 27.71 & \textbf{0.83} & \textbf{0.20} & 3.12 \\
    & + ELBO densification      & 27.73 & 0.83       &0.20      & 2.82 \\
    & + Opacity pruning    & 26.79 & 0.80          & 0.26          & 0.30 \\
    & + Increase iterations         & 26.98 & 0.80          & 0.26          & 0.30 \\
    & + $\mathcal{L}_\text{smt}$    & 27.53 & 0.81          & 0.25          & 0.30 \\
    & Full Model                    & \textbf{27.74} & 0.82 & 0.23 & \textbf{0.30} \\
    \hline
    \multirow{6}{*}{T\&T}
    & 3DGS                          & 24.19 & 0.84          & \textbf{0.19} & 1.57 \\
    & + ELBO densification         & 24.40 & 0.85        &0.20   &0.81  \\
    & + Opacity pruning   & 23.86 & 0.82          & 0.25              & 0.24 \\
    & + Increase iterations         & 24.05 & 0.82          & 0.24          & 0.24 \\
    & + $\mathcal{L}_\text{smt}$    & 24.49 & 0.84          & 0.23          & 0.24 \\
    & Full Model                    & \textbf{24.90} & \textbf{0.85} & 0.21 & \textbf{0.24} \\
    \hline
\end{tabular}
\caption{Ablation study. Evaluate the effectiveness of each component of our model on rendering quality.}
\label{table:ablation1}
\end{table}

\begin{table}[!t]
\small
\centering
\begin{tabular}{@{\hspace{1mm}}l@{\hspace{1mm}}  @{\hspace{1mm}}l@{\hspace{1mm}}@{\hspace{1mm}}l@{\hspace{1mm}} @{\hspace{1mm}}l@{\hspace{1mm}} @{\hspace{1mm}}l@{\hspace{1mm}} @{\hspace{1mm}}l@{\hspace{1mm}}}
    \hline
    Dataset & Method & PSNR$\uparrow$    & SSIM$\uparrow$         & LPIPS$\downarrow$ & $N_{\textrm{GS}}$$\downarrow$ \\
    \hline
    \multirow{4}{*}{Mip-NeRF 360}
      & LightGaussian & 27.35  & 0.82   & 0.23 &  1.07 \\     
      & (+GSDO)       & \textbf{27.75}  & 0.82 & \textbf{0.21} & 1.07  \\
    \cline{2-6}
      & MaskGaussian  & 27.61  & 0.82 & 0.23 & 1.50  \\
      & (+GSDO)       & \textbf{27.94}  & \textbf{0.83} & \textbf{0.20} & 1.50  \\
    \hline
    \multirow{4}{*}{T\&T}
      & LightGaussian & 23.16  & 0.82   & 0.25 & 0.53  \\
      & (+GSDO)       & \textbf{23.97}  & \textbf{0.83} & \textbf{0.22} & 0.53  \\
    \cline{2-6}
      & MaskGaussian  & 24.16  & 0.84 & 0.20 & 0.74   \\
      & (+GSDO)       & \textbf{24.63}  & \textbf{0.85} & \textbf{0.19} & 0.74  \\
    \hline
\end{tabular}
\caption{Integration of GSDO with other 3DGS variants.}
\label{table:ablation3}
 \vspace{-4mm}
\end{table}

\subsection{Generalization of GSDO to Other Methods}
\label{subsec: Generalization to other methods}
Our GSDO module is not restricted to our ADP strategy and can be generalized to other pruning frameworks. Existing pruning-based 3DGS variants can be broadly divided into two categories: mask-guided pruning using scaling or opacity~\cite{lee2024c3dgs,liu2025maskgaussian}, and pruning based on manually defined importance scores~\cite{girish2024eagles,fan2024lightgaussian}. To verify its generalization capability, we integrate the GSDO module as a post-processing component into two representative methods, MaskGaussian and LightGaussian. As shown in~\cref{table:ablation3}, the GSDO module consistently enhances the rendering quality of both methods across the two benchmarks.

\subsection{Efficiency and Robustness Analysis}
\label{subsec: Efficiency and Robustness Analysis}
We evaluate the efficiency of GS\textasciicircum2 in terms of training time and rendering speed. As shown in\cref{tab:train_fps}, integrating GSDO increases training time from 10 to 16 minutes. GS\textasciicircum2 achieves 513 \textit{fps}, surpassing all compared baselines except Mini-Splatting (576 \textit{fps}), while maintaining high rendering quality. As shown in~\cref{tab:mip360_interval}, GS\textasciicircum2 remains robust to the choice of pruning interval. Therefore, we fix the interval to 100 iterations based on experiments, without tuning for dense or sparse scenes.

\begin{table}[t]
\small
\centering
\begin{tabular}{@{}llcc@{}}
\hline
Dataset & Method & \hspace{0.2cm}Train Time (min)$\downarrow$\hspace{0.2cm} & \hspace{0.2cm}FPS$\uparrow$\hspace{0.2cm}  \\
\hline
\multirow{8}{*}{T\&T} 
 & 3DGS & 16 & 135 \\
 & EAGLES & 21 & 247 \\
 & Mini-Splatting & \underline{14} & \textbf{576} \\
 & CompGS & 17 & 325 \\
 & Compact 3DGS & 27 & 345 \\
 & LightGaussian & 17 & 214 \\
 & MaskGaussian & \underline{14} & 353 \\
\cmidrule(lr){2-4}
 & Our (ADP-only) & \textbf{10} & 505 \\
 & Our (Full Model) & 16 & \underline{513} \\
\hline
\end{tabular}
\vspace{2pt}
\caption{Training time and FPS comparison on T \& T.}
\label{tab:train_fps}
\end{table}
\begin{table}[t]
\small
\centering
\begin{tabular}{l lccc}
\hline
Dataset & Interval & PSNR$\uparrow$ &SSIM$\uparrow$&LPIPS$\downarrow$ \\
\hline
\multirow{3}{*}{Mip-NeRF 360}
 & 100 & \textbf{27.74} & \textbf{0.82} & \textbf{0.23} \\
 & 200 & 27.65 & 0.82 & 0.24 \\
 & 400 & 27.60 & 0.82 & 0.24 \\
\hline
\end{tabular}

\caption{Ablation study for different  pruning intervals.}
 \vspace{-4mm}
\label{tab:mip360_interval}

\end{table}

\section{Conclusion}
We propose GS\textasciicircum2, a compact 3D Gaussian Splatting framework that achieves high-quality rendering with significantly fewer Gaussian points. By integrating the Adaptive Densification and Pruning and Graph-based Spatial Distribution Optimization modules, our method enhances the spatial consistency of pruned Gaussian points. Extensive experiments on the Mip-NeRF 360 and Tanks~\&~Temples datasets show that GS\textasciicircum2 outperforms existing methods in both rendering quality and $N_{\textrm{GS}}$, requiring only about 12.5\% of the Gaussian points used in the original 3DGS. Overall, GS\textasciicircum2 improves the accessibility and scalability of high-quality 3D content, making it suitable for real-world applications such as autonomous driving.

\vspace{1mm}\noindent\textbf{Limitations.} While GS\textasciicircum2 effectively reduces the number of Gaussian points when maintaining high rendering quality, it introduces additional training overhead due to the spatial distribution optimization. We leave the reduction of this computational cost for future work.
\section*{Acknowledgment} 
This work is supported by the Fundamental Research Funds for the Central Universities (No. 2025JBZY011) and the National Nature Science Foundation of China (Nos. 62376020 and 62571027). 

{
    \small
    \bibliographystyle{ieeenat_fullname}
    \bibliography{main}

@String(CVPR= {IEEE Conf. Comput. Vis. Pattern Recog.})

@String(ECCV= {Eur. Conf. Comput. Vis.})

@String(TOG= {ACM Trans. Graph.})

@String(TIP  = {IEEE Trans. Image Process.})

@String(ICLR = {Int. Conf. Learn. Represent.})

@String(AAAI = {AAAI})

@String(CVPR  = {CVPR})

@String(ECCV  = {ECCV})

@String(TOG   = {ACM TOG})

@String(TIP   = {IEEE TIP})

@String(ICLR  = {ICLR})

@InProceedings{Chen_2025_CVPR,
    author    = {Chen, Jianchuan and Hu, Jingchuan and Wang, Gaige and Jiang, Zhonghua and Zhou, Tiansong and Chen, Zhiwen and Lv, Chengfei},
    title     = {TaoAvatar: Real-Time Lifelike Full-Body Talking Avatars for Augmented Reality via 3D Gaussian Splatting},
    booktitle = {CVPR},
    year      = {2025},
    pages     = {10723-10734}
}

@inproceedings{ICLR2025_93b4d708,
title={FreeVS: Generative View Synthesis on Free Driving Trajectory},
author={Wang, Qitai and Fan, Lue and Wang, Yuqi and Chen, Yuntao and Zhang, Zhaoxiang},
booktitle={ICLR},
year={2025},
}

@inproceedings{barron2022mip,
  title={Mip-nerf 360: Unbounded anti-aliased neural radiance fields},
  author={Barron, Jonathan T and Mildenhall, Ben and Verbin, Dor and Srinivasan, Pratul P and Hedman, Peter},
  booktitle={CVPR},
  pages={5470--5479},
  year={2022}
}

@inproceedings{mildenhall2021nerf,
  title={NeRF: Representing Scenes as Neural Radiance Fields for View Synthesis},
  author={Mildenhall, Ben and Srinivasan, Pratul P and Tancik, Matthew and Barron, Jonathan T and Ramamoorthi, Ravi and Ng, Ren},
  booktitle={ECCV},
  pages={405--421},
  year={2020}
}

@article{kerbl20233d,
  title={3D Gaussian Splatting for Real-Time Radiance Field Rendering},
  author={Kerbl, Bernhard and Kopanas, Georgios and Leimk{\"u}hler, Thomas and Drettakis, George},
  volume={42},
  number={4},
  pages={139--1},
  journal={TOG},
  year={2023}
}

@inproceedings{lee2024c3dgs,
    author={Lee, Joo Chan and Rho, Daniel and Sun, Xiangyu and Ko, Jong Hwan and Park, Eunbyung},
    title={Compact 3d gaussian representation for radiance field},
    booktitle = {CVPR},
    year      = {2024},
    pages     = {21719-21728}
}

@inproceedings{liu2025maskgaussian,
  title={Maskgaussian: Adaptive 3d gaussian representation from probabilistic masks},
  author={Liu, Yifei and Zhong, Zhihang and Zhan, Yifan and Xu, Sheng and Sun, Xiao},
  booktitle={CVPR},
  pages={681--690},
  year={2025}
}

@inproceedings{girish2024eagles,
  title={Eagles: Efficient accelerated 3d gaussians with lightweight encodings},
  author={Girish, Sharath and Gupta, Kamal and Shrivastava, Abhinav},
  booktitle={ECCV},
  pages={54--71},
  year={2024}
}

@inproceedings{fan2024lightgaussian,
  title={Lightgaussian: Unbounded 3d gaussian compression with 15x reduction and 200+ fps},
  author={Fan, Zhiwen and Wang, Kevin and Wen, Kairun and Zhu, Zehao and Xu, Dejia and Wang, Zhangyang},
 booktitle={NeurIPS},
  volume={37},
  pages={140138--140158},
  year={2024}
}

@article{knapitsch2017tanks,
  title={Tanks and temples: Benchmarking large-scale scene reconstruction},
  author={Knapitsch, Arno and Park, Jaesik and Zhou, Qian-Yi and Koltun, Vladlen},
  journal={TOG},
 volume={36},
  number={4},
  pages={1--13},
  year={2017},
}

@inproceedings{insafutdinov2018unsupervised,
  title={Unsupervised learning of shape and pose with differentiable point clouds},
  author={Insafutdinov, Eldar and Dosovitskiy, Alexey},
  booktitle={NeurIPS},
  volume={31},
  year={2018}
}

@inproceedings{lin2018learning,
  title={Learning efficient point cloud generation for dense 3d object reconstruction},
  author={Lin, Chen-Hsuan and Kong, Chen and Lucey, Simon},
  booktitle={AAAI},
  volume={32},
  number={1},
  year={2018}
}

@article{yifan2019differentiable,
  title={Differentiable surface splatting for point-based geometry processing},
  author={Yifan, Wang and Serena, Felice and Wu, Shihao and {\"O}ztireli, Cengiz and Sorkine-Hornung, Olga},
  journal={TOG},
  volume={38},
  number={6},
  pages={1--14},
  year={2019}
}

@inproceedings{wiles2020synsin,
  title={Synsin: End-to-end view synthesis from a single image},
  author={Wiles, Olivia and Gkioxari, Georgia and Szeliski, Richard and Johnson, Justin},
  booktitle={CVPR},
  pages={7467--7477},
  year={2020}
}

@inproceedings{aliev2020neural,
  title={Neural point-based graphics},
  author={Aliev, Kara-Ali and Sevastopolsky, Artem and Kolos, Maria and Ulyanov, Dmitry and Lempitsky, Victor},
  booktitle={ECCV},
  pages={696--712},
  year={2020}
}

@inproceedings{kopanas2021point,
  title={Point-based neural rendering with per-view optimization},
  author={Kopanas, Georgios and Philip, Julien and Leimk{\"u}hler, Thomas and Drettakis, George},
  booktitle={Computer Graphics Forum},
  volume={40},
  number={4},
  pages={29--43},
  year={2021}
}

@inproceedings{xu2022point,
  title={Point-nerf: Point-based neural radiance fields},
  author={Xu, Qiangeng and Xu, Zexiang and Philip, Julien and Bi, Sai and Shu, Zhixin and Sunkavalli, Kalyan and Neumann, Ulrich},
  booktitle={CVPR},
  pages={5438--5448},
  year={2022}
}

@inproceedings{yu2023mip,
  title={Mip-splatting: Alias-free 3d gaussian splatting},
  author={Yu, Zehao and Chen, Anpei and Huang, Binbin and Sattler, Torsten and Geiger, Andreas},
  booktitle={CVPR},
  pages={19447--19456},
  year={2024}
}

@inproceedings{yan2023multi,
  title={Multi-scale 3d gaussian splatting for anti-aliased rendering},
  author={Yan, Zhiwen and Low, Weng Fei and Chen, Yu and Lee, Gim Hee},
  booktitle={CVPR},
  pages={20923--20931},
  year={2024}
}

@inproceedings{niedermayr2024compressed,
  title={Compressed 3d gaussian splatting for accelerated novel view synthesis},
  author={Niedermayr, Simon and Stumpfegger, Josef and Westermann, R{\"u}diger},
  booktitle={CVPR},
  pages={10349--10358},
  year={2024}
}

@inproceedings{navaneet2024compgs,
  title={Compgs: Smaller and faster gaussian splatting with vector quantization},
  author={Navaneet, KL and Pourahmadi Meibodi, Kossar and Abbasi Koohpayegani, Soroush and Pirsiavash, Hamed},
  booktitle={ECCV},
  pages={330--349},
  year={2024},
}

@inproceedings{morgenstern2023compact,
  title={Compact 3d scene representation via self-organizing gaussian grids},
  author={Morgenstern, Wieland and Barthel, Florian and Hilsmann, Anna and Eisert, Peter},
  booktitle={ECCV},
  pages={18--34},
  year={2024}
}

@inproceedings{lu2024scaffold,
  title={Scaffold-gs: Structured 3d gaussians for view-adaptive rendering},
  author={Lu, Tao and Yu, Mulin and Xu, Linning and Xiangli, Yuanbo and Wang, Limin and Lin, Dahua and Dai, Bo},
  booktitle={CVPR},
  pages={20654--20664},
  year={2024}
}

@inproceedings{yang2024spec,
 title = {Spec-Gaussian: Anisotropic View-Dependent Appearance for 3D Gaussian Splatting},
 author = {Yang, Ziyi and Gao, Xinyu and Sun, Yang-Tian and Huang, Yi-Hua and Lyu, Xiaoyang and Zhou, Wen and Jiao, Shaohui and Qi, Xiaojuan and Jin, Xiaogang},
  booktitle={NeurIPS},
 pages = {61192--61216},
 volume = {37},
 year = {2024}
}

@inproceedings{luiten2023dynamic,
  title={Dynamic 3d gaussians: Tracking by persistent dynamic view synthesis},
  author={Luiten, Jonathon and Kopanas, Georgios and Leibe, Bastian and Ramanan, Deva},
  booktitle={3DV},
  pages={800--809},
  year={2024},
}

@inproceedings{yang2023deformable,
  title={Deformable 3d gaussians for high-fidelity monocular dynamic scene reconstruction},
  author={Yang, Ziyi and Gao, Xinyu and Zhou, Wen and Jiao, Shaohui and Zhang, Yuqing and Jin, Xiaogang},
  booktitle={CVPR},
  pages={20331--20341},
  year={2024}
}

@inproceedings{wu20244d,
   title={4d gaussian splatting for real-time dynamic scene rendering},
  author={Wu, Guanjun and Yi, Taoran and Fang, Jiemin and Xie, Lingxi and Zhang, Xiaopeng and Wei, Wei and Liu, Wenyu and Tian, Qi and Wang, Xinggang},
  booktitle={CVPR},
  pages={20310--20320},
  year={2024}
}

@inproceedings{yang2023real,
  title={Real-time Photorealistic Dynamic Scene Representation and Rendering with 4D Gaussian Splatting},
  author={Yang, Zeyu and Yang, Hongye and Pan, Zijie and Zhang, Li},
  booktitle={ICLR},
  year={2024}
}

@inproceedings{katsumata2023efficient,
  title={A compact dynamic 3d gaussian representation for real-time dynamic view synthesis},
  author={Katsumata, Kai and Vo, Duc Minh and Nakayama, Hideki},
  booktitle={ECCV},
  pages={394--412},
  year={2024}
}

@inproceedings{kratimenos2023dynmf,
  title={Dynmf: Neural motion factorization for real-time dynamic view synthesis with 3d gaussian splatting},
  author={Kratimenos, Agelos and Lei, Jiahui and Daniilidis, Kostas},
  booktitle={ECCV},
  pages={252--269},
  year={2024}
}

@inproceedings{huang2023sc,
  title={Sc-gs: Sparse-controlled gaussian splatting for editable dynamic scenes},
  author={Huang, Yi-Hua and Sun, Yang-Tian and Yang, Ziyi and Lyu, Xiaoyang and Cao, Yan-Pei and Qi, Xiaojuan},
  booktitle={CVPR},
  pages={4220--4230},
  year={2024}
}

@inproceedings{zhang2024lp,
  title={Lp-3dgs: Learning to prune 3d gaussian splatting},
  author={Zhang, Zhaoliang and Song, Tianchen and Lee, Yongjae and Yang, Li and Peng, Cheng and Chellappa, Rama and Fan, Deliang},
  booktitle={NeurIPS},
  volume={37},
  pages={122434--122457},
  year={2024}
}

@inproceedings{PUP3DGS,
  title={Pup 3d-gs: Principled uncertainty pruning for 3d gaussian splatting},
  author={Hanson, Alex and Tu, Allen and Singla, Vasu and Jayawardhana, Mayuka and Zwicker, Matthias and Goldstein, Tom},
  booktitle={CVPR},
  pages={5949--5958},
  year={2025}
}

@inproceedings{chen2024hac,
  title={Hac: Hash-grid assisted context for 3d gaussian splatting compression},
  author={Chen, Yihang and Wu, Qianyi and Lin, Weiyao and Harandi, Mehrtash and Cai, Jianfei},
  booktitle={ECCV},
  pages={422--438},
  year={2024}
}

@inproceedings{fang2024mini,
  title={Mini-splatting: Representing scenes with a constrained number of gaussians},
  author={Fang, Guangchi and Wang, Bing},
  booktitle={ECCV},
  pages={165--181},
  year={2024}
}

@article{kingma2013auto,
  title={Auto-encoding variational bayes},
  author={Kingma, Diederik P and Welling, Max},
  journal={arXiv preprint arXiv:1312.6114},
  year={2013}
}

@article{blei2017variational,
  title={Variational inference: A review for statisticians},
  author={Blei, David M and Kucukelbir, Alp and McAuliffe, Jon D},
  journal={Journal of the American statistical Association},
  volume={112},
  number={518},
  pages={859--877},
  year={2017},
}

@article{wang2004image,
  title={Image quality assessment: from error visibility to structural similarity},
  author={Wang, Zhou and Bovik, Alan C and Sheikh, Hamid R and Simoncelli, Eero P},
  journal={TIP},
  volume={13},
  number={4},
  pages={600--612},
  year={2004}
}

@InProceedings{10.1007/978-3-031-73397-0_10,
  title={Gaussian grouping: Segment and edit anything in 3d scenes},
  author={Ye, Mingqiao and Danelljan, Martin and Yu, Fisher and Ke, Lei},
  booktitle={ECCV},
  pages={162--179},
  year={2024}
}

@inproceedings{zhang2018perceptual,
  title={The unreasonable effectiveness of deep features as a perceptual metric},
  author={Zhang, Richard and Isola, Phillip and Efros, Alexei A and Shechtman, Eli and Wang, Oliver},
  booktitle={CVPR},
  pages={586--595},
  year={2018}
}

@inproceedings{niemeyer2024radsplat,
  title={Radsplat: Radiance field-informed gaussian splatting for robust real-time rendering with 900+ fps},
  author={Niemeyer, Michael and Manhardt, Fabian and Rakotosaona, Marie-Julie and Oechsle, Michael and Duckworth, Daniel and Gosula, Rama and Tateno, Keisuke and Bates, John and Kaeser, Dominik and Tombari, Federico},
  booktitle={3DV},
  pages={134--144},
  year={2025}
}
}

\end{document}